\documentclass[journal]{IEEEtran}

\usepackage[cmex10]{amsmath}
\usepackage{siunitx}
\usepackage[font=footnotesize]{caption}
\usepackage{psfrag}
\usepackage[utf8]{inputenc}
\usepackage[T1]{fontenc}
\usepackage{amsmath,amsfonts,amsbsy,amssymb}
\usepackage{mathabx}
\usepackage{mathrsfs}
\usepackage[nolist]{acronym}
\usepackage{tabularx}
\usepackage{amssymb}
\usepackage{pdfpages}
\usepackage{amsmath}
\usepackage{graphicx}
\usepackage{soul}
\usepackage{cite}
\usepackage{wasysym}
\usepackage{float}
\usepackage{xcolor}
\usepackage{subfigure}
\usepackage{ragged2e}
\usepackage{makecell, multirow, tabularx}
\usepackage{booktabs}

\usepackage{siunitx, mhchem}
\usepackage{xcolor}
 
\usepackage[ruled, lined, longend, linesnumbered]{algorithm2e}

\usepackage{xcolor}
\usepackage{url}
\usepackage{multicol}
\usepackage{verbatim}
\usepackage{array}
\definecolor{uranianblue}{rgb}{0.69,0.86,0.96}
\definecolor{thistle}{rgb}{0.85,0.75,0.85}
\definecolor{mintgreen}{rgb}{0.6,1.0,0.6}
\definecolor{new}{rgb}{0.4940 0.1840 0.5560}
\definecolor{azurexwebcolor}{rgb}{0.94,1.0,1.0}

\newcommand\cparagraph[1]{\vspace{0.6mm}\noindent\textbf{#1.}}

\begin{document}
\title{Coordinated Multi-Drone Last-mile Delivery: Learning Strategies for Energy-aware \\ and Timely Operations}

\author{\IEEEauthorblockN{Chuhao Qin\IEEEauthorrefmark{1}, Arun Narayanan\IEEEauthorrefmark{2} and 
Evangelos Pournaras\IEEEauthorrefmark{1}} \\

\IEEEauthorblockA{
\IEEEauthorrefmark{1}School of Computer Science, University of Leeds, UK\\
\IEEEauthorrefmark{2}School of Energy Systems, LUT University, Finland
}
\thanks{}
}


\maketitle

\begin{abstract}
Drones have recently emerged as a faster, safer, and cost-efficient way for last-mile deliveries of parcels, particularly for urgent medical deliveries highlighted during the pandemic. This paper addresses a new challenge of multi-parcel delivery with a swarm of energy-aware drones, accounting for time-sensitive customer requirements. Each drone plans an optimal multi-parcel route within its battery-restricted flight range to minimize delivery delays and reduce energy consumption. The problem is tackled by decomposing it into three sub-problems: (1) optimizing depot locations and service areas using K-means clustering; (2) determining the optimal flight range for drones through reinforcement learning; and (3) planning and selecting multi-parcel delivery routes via a new optimized plan selection approach. To integrate these solutions and enhance long-term efficiency, we propose a novel algorithm leveraging actor-critic-based multi-agent deep reinforcement learning. Extensive experimentation using realistic delivery datasets demonstrate an exceptional performance of the proposed algorithm. We provide new insights into economic efficiency (minimize energy consumption), rapid operations (reduce delivery delays and overall execution time), and strategic guidance on depot deployment for practical logistics applications.

\end{abstract}


\IEEEpeerreviewmaketitle

\section{Introduction \label{sec:intro}}
Unmanned aerial vehicles (UAVs), commonly known as drones, have gained significant attention as a solution for last-mile delivery, especially in recent years~\cite{kitjacharoenchai2020two}. For instance, the COVID-19 pandemic has highlighted the vulnerabilities of traditional delivery methods, as deliverymen risk spreading the virus. This was particularly problematic in quarantine zones, where customers faced difficulties in accessing logistics services~\cite{brooks2020psychological,khan2020robotics}. In contrast, drones offer a safer and more flexible alternative. Due to their high mobility, carrying capacity, and accurate GPS navigation, drones are able to deliver parcels directly to small places such as doorways and balconies, avoiding human contact and traffic congestion. Furthermore, drones help alleviate the workload of ground vehicles, such as vans and trucks, thereby reducing road congestion. This shift from ground-based to aerial operations contributes to lower fuel consumption and emissions, supporting broader efforts toward achieving net zero targets~\cite{bukhari2023zero}. As a consequence, major companies such as Amazon, Meituan, and Zipline have been working on delivery systems that use a swarm of drones, aiming to complete delivery missions while reducing costs~\cite{ackerman2019blood}.

Route planning is vital to enhance the efficiency of drone swarms in completing delivery tasks~\cite{dorling2016vehicle}. In coordinated delivery, effective route planning guides swarms of drones to reach customers in the shortest possible time or distance. Prior works model this routing problem, which is essentially a variant of the Traveling Salesman Problem (TSP) and Vehicle Routing Problem (VRP), as a Mixed-Integer Programming (MIP) problem and uses various heuristic algorithms to solve the complex combinatorial optimization; approaches such as genetic algorithms, simulated annealing, ant colony optimization, and reinforcement learning have been widely applied~\cite{hong2023logistics,wu2021reinforcement,pournaras2018decentralized}. However, existing models for routing planning of drone delivery still faces two technical challenges.

The first challenge is \textit{energy-aware planning}. Each drone must ensure its battery level remains above a minimum safety threshold (typically $25-30\%$) to maintain operational longevity and avoid mission failures. Moreover, energy consumption directly affects carbon emissions, which are increasingly important for achieving net zero goals~\cite{bukhari2023zero}. While modern drones are capable of multi-parcel deliveries, existing models that solely optimize delivery time are insufficient, since they fail to account for the power consumption variation caused by parcel weight~\cite{stolaroff2018energy,qin20223,10639825}. Heavier payloads lead to significantly higher energy cost and need to deliver first if the requested customers are close. Therefore, drones must be aware of their full delivery route and select energy-optimal plans. Unlike traditional models, we use an energy consumption model~\cite{stolaroff2018energy} that requires drones to have full knowledge of their entire route within a specific time window, allowing them to carry all parcels initially and calculate energy consumption accurately. 

The second challenge is \textit{delivery delay}. Delivery delay becomes a key concern when drone resources are limited and not all customer requests can be fulfilled within the maximum flight time of drones. This is especially critical in time-sensitive scenarios, such as delivering medical supplies during pandemics~\cite{brooks2020psychological,gentili2022locating}. While some existing approaches address this by prioritizing requests with the highest delivery delay (i.e., the actual arrival time minus the expected deadline)~\cite{dukkanci2021minimizing}, it remains challenging to determine which requests to prioritize in real time. This is because customers' priority changes with time, there is a new set of customers who must be serviced at every time step. Drones know the current distribution and delivery time of customers, but do not know when and where the customers will appear in a subsequent time step. As a result, drones have to adapt to an unexpected environment in terms of the number and locations of customers and their demands. Motivated by this, we leverage the deep reinforcement learning that assists drones to anticipate areas of high future delay and proactively position themselves, even if that means temporarily overlooking requests that are already delayed. This ``slower is faster'' strategy helps to minimize overall delay across the entire delivery horizon. 


Therefore, this paper introduces the Delay-and-Energy-aware Drone Delivery Problem (DEDDP), a novel route planning problem that uses a swarm of drones to deliver packages. The main contributions are outlined as follows: (1) We model the DEDDP as an MIP that aims to minimize both delivery delay and energy consumption of drones while respecting their own battery and payload capacity. To the best of our knowledge, this is the first attempt to study mulit-drone delivery problem accounting for both time-sensitive customer needs and energy consumption of multi-parcel route planning. (2) The DEDDP is divided into three sub-problems. First, the \textit{customer segmentation} problem partitions the map into service areas based on customer distribution and sets their centers as depots. Second, the \textit{flight range selection} problem assigns a specific flight range to each drone based on departure/destination depots for long-term efficiency of routing planning. Third, the \textit{route planning} problem finds the optimal delivery routes for drones within their assigned flight ranges. (3) We propose a novel algorithm, named Multi-Agent Reinforcement learning-based Optimized Plan Selection (\emph{MAR-OPS})\footnote{The code is available as open source code on github at: https://github.com/TDI-Lab/MAR-OPS.}, that integrates K-means clustering, DRL, and a MIP solver (e.g., Gurobi~\cite{gurobi_2024}) to solve the three subproblems, respectively. Furthermore, \emph{MAR-OPS} leverages the state-of-the-art DRL techniques, such as actor-critic networks and proximal policy optimization, to improve the training stability. (4) Quantitative findings using real-world data and emulated last-mile delivery scenarios validate the superior performance of the proposed approach over baseline methods. It provides new insights into a sustainable (low energy consumption and carbon emission), timely (low delivery delays), and adaptive (high operation speed) multi-drone delivery.

\section{Related Work} \label{sec:related_work}

\begin{table*}[ht]
\footnotesize
\centering
\caption{Comparison to related work; criteria covered (\checkmark) or not ($\times$). Here, CL: Collective Learning; DRL: Deep Reinforcement Learning; Heur.: heuristics; Opt. Optimization; MDP: Markov Decision Process}
\label{tab:comparison}
\renewcommand{\arraystretch}{1}
\begin{tabular}{lccccccccc}
\toprule
\textbf{Criteria} \;  \textbf{Ref.:}& 
\textcolor{blue}{\cite{luo2021multi}} & 
\textcolor{blue}{\cite{wang2023multi}} &
\textcolor{blue}{\cite{bi2024truck}} & 
\textcolor{blue}{\cite{deng2024stochastic}} & 
\textcolor{blue}{\cite{liu2024cooperated}} & 
\textcolor{blue}{\cite{zhang2025drone}} & 
\textcolor{blue}{\cite{narayanan2024large}} &
\textbf{This work} \\
\midrule
\textbf{No. of Customers} & $25$--$100$ & $0$--$50$ & $15$--$36$ & $10$--$25$  & $10$--$100$ & $5$--$50$ & $10$--$10000$ & $400-600$ \\
\textbf{Method} & Opt., Heur. & Non-linear Opt. & DRL & MDP & Opt., Heur. & Opt., Heur. & Heur., CL & Heur., DRL \\
\textbf{Multi-visit} & \checkmark & $\times$ &  $\times$ & $\times$ & \checkmark & $\times$ & \checkmark & \checkmark \\
\textbf{Energy awareness} & \checkmark & $\times$  & $\times$ & \checkmark & \checkmark & \checkmark & \checkmark & \checkmark \\
\textbf{Delay awareness} & $\times$ & \checkmark &  $\times$ & $\times$ & \checkmark & $\times$ & $\times$ & \checkmark \\
\textbf{Scalability to large-scale systems} & $\times$ & $\times$  & $\times$ & $\times$ & $\times$ & $\times$ & \checkmark & \checkmark \\
\textbf{Adaptability to dynamic environments} & \checkmark & \checkmark  & $\times$ & $\times$ & \checkmark & $\times$ & $\times$ & \checkmark \\
\textbf{Sustainable long-term efficiency} & \checkmark & $\times$ & $\times$ & $\times$ & \checkmark & \checkmark & $\times$ & \checkmark \\
\bottomrule
\end{tabular}
\end{table*}


The proposed work differs from the existing literature in the following aspects. First, existing methods to overcome the limitations of the range and payload capacity of drones have been to pair them with delivery trucks such that drones only perform deliveries within small ranges (``last-mile deliveries'') \cite{murray2015flying, chung2020optimization,freitas2023exact,stodola2024multi}. Furthermore, the use of multiple depots \cite{ham2018integrated,stodola2024multi} and recharging (in-flight or with distributed charging stations) \cite{song2018persistent,liu2024optimal} has been proposed previously. However, synchronization between drones and trucks is a critical and complex problem with significant costs due to idle times \cite{freitas2023exact}. These solutions are also not environmentally friendly; drones typically have fewer CO$_{2}$ emissions than delivery trucks, and life-cycle impacts are increased if additional depots are required \cite{stolaroff2018energy,chiang2019impact}. Moreover, drone deliveries without trucks become cost-competitive when more packages can be carried per tour \cite{choi2021comparison}. Only a few models have explored swarm-based \textit{multi-visit drone delivery} scenarios where numerous drones operate together to make \textit{multiple} deliveries directly from the depot \cite{chung2020optimization,madani2022hybrid,narayanan2024large}, or from a truck \cite{luo2021multi}. Our work differs from the dominant truck--drone models by studying a more  universal scenario where drones carry multiple packages directly from a depot and deliver packages to customers based on the urgency of their delivery.

Second, we perform \textit{delay-aware drone deliveries} in which a customer's delivery priority increases with the waiting time; in other words, it is more urgent to deliver packages to customers who have had to wait longer for the deliveries. In contrast, most of the literature so far has considered priorities that remain constant through the entire service period \cite{pasha2022drone,wang2023multi, narayanan2024large}. Wang et al. \cite{wang2023multi}, for example, employed fuzzy theory to deal with demand arrival rate under a priority queuing strategy, but their priority queue remained constant throughout the operational period. Similarly, Narayanan et al. \cite{narayanan2024large} considered service delays by using a system of prioritization to compensate the customers differently for late deliveries; however, they also used a fixed priority queue.

Third, most of the papers in the literature consider energy consumption as a fixed limit on drone flight time or flight range \cite{chung2020optimization,zhang_energy_2021}. It is only recently that some related research has considered the battery energy consumption rate \cite{di2021last,narayanan2024large,zhang2025drone}. In this paper, we consider the battery energy consumption rate with multiple package deliveries. When a package is delivered, the weight of the drone decreases, thereby reducing the energy consumption rate and increasing the range. Unlike most prior literature that consider a single package and use a direct relationship between consumption and weight, we explicitly account for this increase with the number of packages. Furthermore, very few studies have considered both delay awareness and energy consumption together. Some studies have examined the tradeoff between energy consumption and delivery time (e.g., \cite{liu2024cooperated,bi2024truck}), but they consider fixed delivery times, essentially applying the capacitated vehicle routing problem with time windows to drone deliveries.

Finally, the proposed approach is different from most drone delivery systems that typically consider static environments; for example, the previous work \cite{wu2019deepeta} proposed DeepETA, a spatial-temporal sequential neural network model for estimating travel (arrival) time for parcel delivery, formulated with multiple destinations. However, this approach relies on the availability of information regarding delivery routes. In Deng at al. \cite{deng2024stochastic}, the authors considered travel time uncertainty and the minimization of service delay risks (with a truck-and-drone delivery system), but the drones had complete information. Recent research that has employed reinforcement learning approaches, e.g., \cite{shu2024energy,suanpang2024optimizing,tarhan2024genetic} focused on relearning flight paths and delivery strategies, assuming complete information of all customers and possible routes. In contrast, our approach is adaptive to new customers and locations that change with increasing demand urgency at different times. It uses DRL to improve optimization by observing new delivery demand and take strategic policies to visit the customers with the most urgent priority. This makes our approach highly adaptive to various operational scenarios and conditions. 

Table \ref{tab:comparison} compares the most prominent recent literature. Here, \textit{multi-visit} refers to whether a drone can carry multiple packages; \textit{energy-awareness} implies that energy consumption is considered; \textit{delay-awareness} checks if delivery delays have been considered; \textit{scalability to large-scale systems} refers to the ability of methods to handle increasing numbers of customers, drones, tasks, and environmental complexity; \textit{adaptability to dynamic environments} refers to the responsiveness of methods to real-time and uncertain changes, such as the distribution of customer requests; and \textit{sustainable long-term efficiency} refers to the capacity of methods to optimize resource usage, such as energy and time, over extended periods.

\begin{table}  
    \footnotesize
	\caption{Notations.}  
	\centering
	\begin{tabularx}{\linewidth}{lXl}  
		\hline  
		Notation & Explanation \\  
		\hline     
		$ u, U, \mathcal{U} $  & Index of a drone; total number of drones; set of drones \\ 
        $ n, N, \mathcal{N} $  & Index of a depot; total number of depots; set of depots \\
        $ \mathcal{V}, \mathcal{C} $  & Set of nodes; set of customer nodes \\
        $ t, T, \mathcal{T} $  & Index of a time window; total number of time windows; set of time windows \\ 
        $ t_i $  & The expected delivery time window demanded by the customer request $i$ \\
        $ s_i $  & The delivery status of customer request $i$ \\
        $ f_p $ & Function of delivery delay \\
        $ f_u $ & Function of energy consumption of $u$ \\
        $ m^{\mathsf{body}}_u $ & Body mass of drone \\
        $m^{\mathsf{battery}}_u$ & battery mass of drone \\
        $ v, \hat{v} $ & Ground speed of drone; the induced velocity \\
        $ d, r $ & Diameter of propellers; Number of propellers \\
        $ \theta, \epsilon_u $ & The pitch angle of drone; the power efficiency of drone \\
        $ x_{uij} $  & Binary variable which takes $1$ if drone $u$ travels from node $j$ to node $i$ and $0$ otherwise. \\
        $ d_{ij} $  & Traveling distance between node $i$ and node $j$ \\
        $ E_{uij} $  & Energy consumption of $u$ when it travels from node $j$ to $i$ \\
        $ m^{\mathsf{parcel}}_{uij} $  & The weight of parcels carried by drone $u$ from node $j$ to $i$ \\
        $ M_u, R_u $  & The maximum payload of $u$; maximum flight range of $u$ \\
        $ \alpha $ & The parameter weight to balance the trade-off between delivery delay and energy consumption  \\
		\hline  
	\end{tabularx}  
	\label{table:notation}
\end{table}

\begin{figure}
    \centering
    \includegraphics[width=\linewidth]{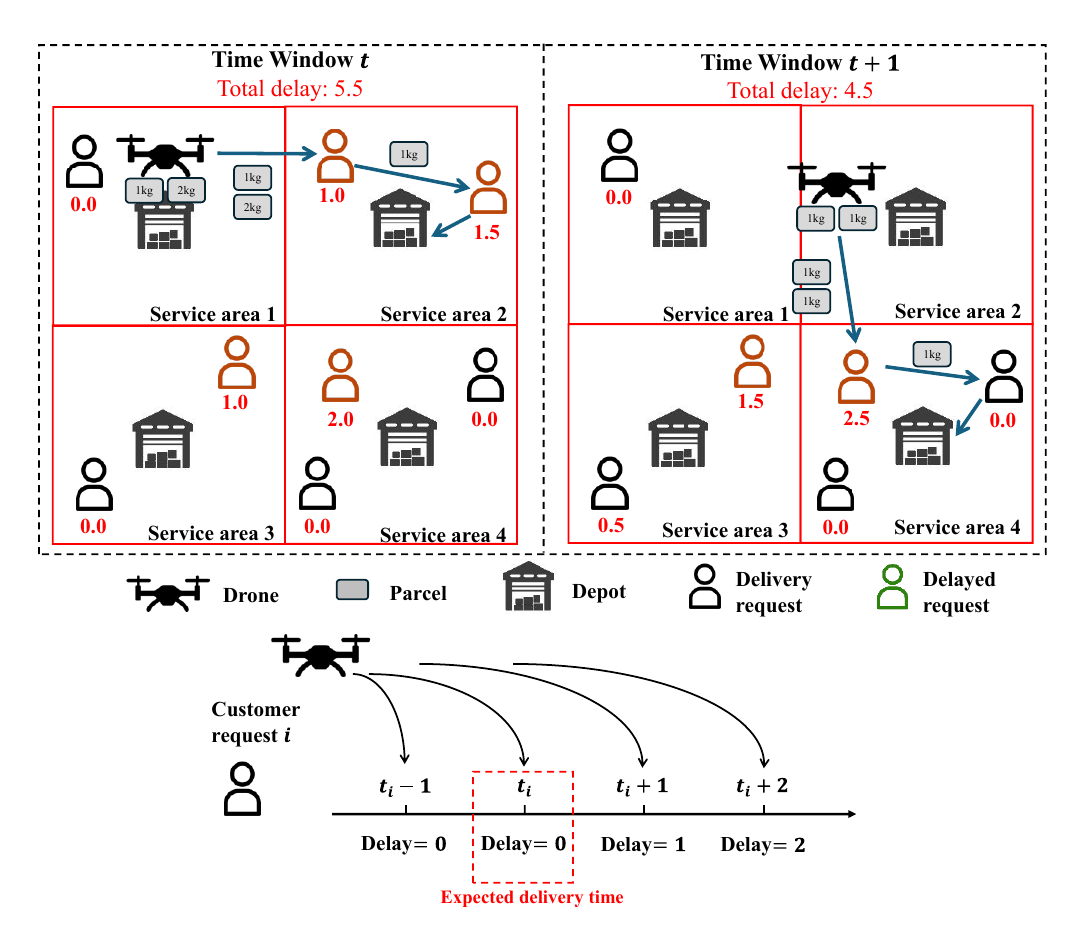}
    \caption{The scenario of drone delivery over three service areas within two consecutive time windows. The blue dashed arrows represent the traveling route of the drone. The red values represent the delivery delay of each customer.}
    \label{fig:scenario}
\end{figure}

\section{System Model \label{sec:model}}
In this section, we define key concepts of scenarios and then formulate the main performance metrics. Table~\ref{table:notation} illustrates the list of mathematical notations used in our model.

\subsection{Definitions and assumptions}
Consider a swarm of drones $\mathcal{U} \triangleq \{1,2,...,U\}$ taking parcels from depots and delivering them to customer requests over a 2D map within a set of specified time windows $\mathcal{T} \triangleq \{1,2,...,T\}$. The whole map is partitioned into a number of service areas, each denoting a geographically bounded region, within which a drone can efficiently serve customer requests. Then, a delivery center, i.e., the depot, is set up in the center of each service area. Both depots and customer requests are the nodes distributed in the map, denoted as $\mathcal{V}$. At the time window $t$, the nodes can be defined as $\mathcal{V}(t) = \mathcal{N} \cup \mathcal{C}(t)$, where $\mathcal{N}$ denotes the set of depots, $\mathcal{N} \triangleq \{1,2,...,N\}$, and $\mathcal{C}(t)$ indicates the set of customer nodes within time window $t$. 

To represent the traveling information of drones, the model uses the binary variable $x_{uij}$, which takes 1 if drone $u$ travels from node $j$ to node $i$ and 0 otherwise. At each time window, drones execute the multi-parcel delivery, that is, each of them departs from a depot, deliver parcels to multiple customer requests, and returns back to the original depot or other depots that locate at the neighboring service areas. Therefore, the node $j$ comes from the set of depots $j \in \mathcal{N}$, at the beginning of a trip, and becomes the node of customer request $j \in \mathcal{C}(t)$ after that. In contrast, the node $i$ is the depot node $i \in \mathcal{N}$ at the end of a trip.

The set of customer requests is updated at every time window by excluding the nodes of delivered requests and including new requests in the next time window. Thus, the updating of customer requests nodes from $t$ to $t + 1$ is formulated as follows:
\begin{align}
    \begin{split}
    \mathcal{C}(t+1) = \Big( \mathcal{C}(t) \backslash	
    & \{ k \in \mathcal{C}(t) | s_i = 0 \} \Big) \\
    & \cup \Big( k \in \mathcal{C}(t_i) | t_i = t + 1 \Big),
    \end{split}
\end{align}
where $\mathcal{C}(t)$ indicates the set of customer nodes existed within time window $t$, $t \in \mathcal{T}$; $k$ is the index of a customer node; $t_i$ denotes the delivery time window demanded by customer request $i$; and $s_i$ indicates the delivery status of request $i$, which takes $0$ if the customer is delivered by a drone and $1$ otherwise. It can be expressed as follows:
\begin{equation}
    s_i = 1 - \min(\sum_{t=1}^t \sum_{u \in \mathcal{U}} \sum_{j \in \mathcal{V}(t)} x_{uij}(t), 1),
\end{equation}
where $\mathcal{V}(t)$ is the set of all nodes in the map within time window $t$, and $\mathcal{V}(t) = \mathcal{C}(t) \cup \mathcal{N}$, including both customer nodes and depot nodes; $x_{uij}$ is the binary variable which takes 1 if drone $u$ travels from node $j$ to customer node $i$ and 0 otherwise.

The delivery delay is also defined to measure the amount of time by which a drone arrive later than the expected delivery time $t_i$ of a customer request, i.e., $t > t_i$. Meanwhile, if the request $i$ is delivered ($s_i = 0$), its delivery delay is set as $0$ since it is removed from the map. The model calculates the delivery delay $f_p(\cdot)$, which is formulated as follows:
\begin{equation}
    \begin{split}
        f_p(t, t_i) = \left \{
        \begin{array}{ll}
           (t - t_i) \cdot s_c,       &    t_i \leq t, t_i \in \mathcal{T} \\
           0,       &    otherwise
        \end{array}.
        \right.
    \end{split}
\end{equation}
The delivery delay prioritizes timely deliveries by penalizing late arrivals, ensuring more urgent requests are met promptly. This can lead to higher customer satisfaction and retention, ultimately enhancing long-term quality of service.

Therefore, as shown in Fig.~\ref{fig:scenario}, the overall operational flow is listed as follows: In a time window $t$, each drone observes the delivery delay of undelivered requests. The delay is set as $0$ at $t = 0$ but increases if the current time $t$ is higher than the expected delivery time $t_i$ of request, $c \in \mathcal{C}(t)$. Once the drone determines which requests it serves, it takes all required parcels initially and unloads them to these customer requests one by one. Then, the drone returns to the nearest depot where it swaps a new battery for charging and continues observing the status of customer requests at $t+1$. The delivered requests are removed from the map, whereas the undelivered ones increase their delivery delay. After that, the drone pick-ups parcels and continues delivering at the next time window.  

Furthermore, several necessary and realistic assumptions are made as follows: (1) Each drone flies at a constant ground speed and has a fixed battery capacity that limits the flight range; (2) Each customer is visited only once and by only one drone; (3) Customers who request parcels within the same time window have the same delivery delay; (4) The time of swapping batteries of drones (i.e., recharging), taking-off/landing, and loading parcels is not considered in this paper. (5) A time window is assumed to be longer than the maximum flight time of drones. Note that this general model is also well-suited for supporting truck-drone operations: trucks act as ``mobile depots'', traveling to the centers of service areas while carrying parcels and batteries for drone delivery. In addition, trucks can observe customer demand and adjust their routes by moving to other service areas as needed.

\subsection{Energy Consumption Model} \label{sec:energy_model}
Energy consumption for delivery indicates the energy required by a drone to fly through a route. Drones spend energy to surpass gravity force and counter drag forces due to wind and forward motions. A drone controls the speed of each rotor to achieve the thrust and pitch necessary to hover and travel forward at the desired velocity while balancing the drone payload and drag forces~\cite{stolaroff2018energy}. For a drone $u$ with mass $m^{\mathsf{body}}_u$ and its battery with mass $m^{\mathsf{battery}}_u$, we define the total required thrust as follows:
\begin{equation}
	\mathcal{T}_{uij} = m_{uij} \cdot g \cdot (1 + tan(\theta)),
\end{equation}
\begin{equation}
    m_{uij} = m^{\mathsf{body}}_u + m^{\mathsf{battery}}_u + m^{\mathsf{parcel}}_{uij},
\end{equation}
\begin{equation}
    m^{\mathsf{parcel}}_{uij} = m^{\mathsf{parcel}}_{ujk} - m^{\mathsf{parcel}}_{j},
    \label{eq:weight_update}
\end{equation}
\noindent where $g$ is the gravitational constant; $\theta$ is the pitch angle that depends on air speed and air density; and $m^{\mathsf{parcel}}_{uij}$ is the weight of parcels carried in drone $u$ from node $j$ to node $i$, where $i, j \in \mathcal{V}(t)$. Eq.(\ref{eq:weight_update}) implies that the drone $u$ starts by carrying all assigned parcels and uploads each one to the corresponding customer sequentially, $k \rightarrow j \rightarrow i$, $\forall i,j,k \in \mathcal{V}(t)$. Note that drones carry no parcel to depots, i.e., $m^{\mathsf{parcel}}_{j} = 0$. Given a thrust $\mathcal{T}_{uij}$, the induced velocity can be found by solving the nonlinear equation~\cite{stolaroff2018energy}:
\begin{equation}
	\hat{v} = \frac{2 \cdot \mathcal{T}_{uij}}{\pi \cdot d^{2} \cdot r \cdot \rho \cdot \sqrt{(v \cdot cos\theta)^{2} + (v \cdot sin\theta + \hat{v})^{2}}},
\end{equation} 
\noindent where $v$ is the average ground speed; $d$ and $r$ are the diameter and number of drone rotors; $\rho$ is the density of air. Therefore, given the overall power efficiency $\epsilon_u$, the power consumption with forward velocity and forward pitch is formulated as~\cite{stolaroff2018energy}:
\begin{equation}
    P_{uij} = (v \cdot sin\theta + \hat{v}) \cdot \frac{\mathcal{T}_{uij}}{\epsilon_u}.
	\label{eq:power_consumption}
\end{equation}
The power consumption can be formulated as $P_{uij} = f_{u}(m^{\mathsf{parcel}}_{uij})$ to represent its relationship with the weight of parcels carried by $u$. The energy consumption of $u$ that travels the distance $d_{ij}$ from node $j$ to node $i$ is given by:
\begin{equation}
    E_{uij} = \frac{P_{uij} \cdot d_{ij}}{ v } = f_{u}(m^{\mathsf{parcel}}_{uij}) \cdot \frac{d_{uij}}{v}.
    \label{eq:energy_consumption}
\end{equation}

\subsection{Problem Formulation}
Based on the defined scenarios settings and assumptions, our DEDDP can be stated as: \textit{Find optimal routes of a swarm of drones $\mathcal{U}$ over all time windows $\mathcal{T}$ such that the customers $\mathcal{C}$ with different delivery time demands receive parcels with minimum time delay while minimizing the energy consumption of drones.} Each route of drone $u$ at time window $t$, which can be represented by $x_{uij}$, $i,j \in \mathcal{V}$, starts and ends at depots, passing through several customers. The time delay and energy consumption in DEDDP are calculated based on the defined models.

Our DEDDP model can be formulated as follows:
\begin{equation}
    \min \; \sum_{t \in \mathcal{T}} \sum_{i \in \mathcal{V}(t)} f_p (t, t_i),
    \label{eq:objective_delay}
\end{equation}
\begin{equation}
    \min \; \sum_{t \in \mathcal{T}} \sum_{u \in \mathcal{U}} \sum_{i \in \mathcal{V}(t)} \sum_{j \in \mathcal{V}(t)} E_{uij} \cdot x_{uij} (t),
    \label{eq:objective_energy}
\end{equation}

\noindent Subject to
\begin{equation}
    \sum_{t \in \mathcal{T}} \sum_{u \in \mathcal{U}} \sum_{j \in \mathcal{V}(t)} x_{uij} (t) = 1, \forall i \in \mathcal{C}(t),
    \label{eq:constraints_customer}
\end{equation}
\begin{equation}
    \sum_{i \in \mathcal{N}} \sum_{j \in \mathcal{V}(t)} x_{uij} (t) \leq 1, \forall u \in \mathcal{U}, \forall t \in \mathcal{T},
    \label{eq:constraint_depot}
\end{equation}
\begin{equation}
    \sum_{j \in \mathcal{V}(t)} x_{uij} (t) = \sum_{j \in \mathcal{V}(t)} x_{uji} (t), \forall i \in \mathcal{V}(t), \forall u \in \mathcal{U}, \forall t \in \mathcal{T},
    \label{eq:constraint_flow}
\end{equation}
\begin{equation}
    \sum_{i \in \mathcal{V}(t)} \sum_{j \in \mathcal{V}(t)} d_{ij} \cdot x_{uij} (t) \leq R_u,
    \label{eq:constraint_battery}
\end{equation}
\begin{equation}
    m^{\mathsf{parcel}}_{uij} \leq M_u.
    \label{eq:constraint_weight}
\end{equation}

The objective function (\ref{eq:objective_delay}) aims to minimize the delivery delay of customers. The objective function (\ref{eq:objective_energy}) aims to minimize the total energy cost of drones. Constraint (\ref{eq:constraints_customer}) presents that every customer is visited exactly once. Constraint (\ref{eq:constraint_depot}) ensures that all routes must start and end at depots. Constraint (\ref{eq:constraint_flow}) is a connectivity constraint to ensure that each drone leaves each customer after delivery. Constraint (\ref{eq:constraint_battery}) restricts that the maximum flight range of drone is within $R_u$, which covers the whole map as default. Constraint (\ref{eq:constraint_weight}) restricts the total weight of parcels carried in the drone to be always within its maximum payload $M_u$.

\section{Methodology \label{sec:methods}}

The DEDDP is decomposed into the following three sub-problems and the solutions are then combined: a \textit{customer requests segmentation} problem, a \textit{flight range selection} problem and a \textit{routing planning} problem of drones within the flight range. In this section, the novel methodologies for each of these sub-problems are introduced.

\begin{figure}
    \centering
    \includegraphics[width=\linewidth]{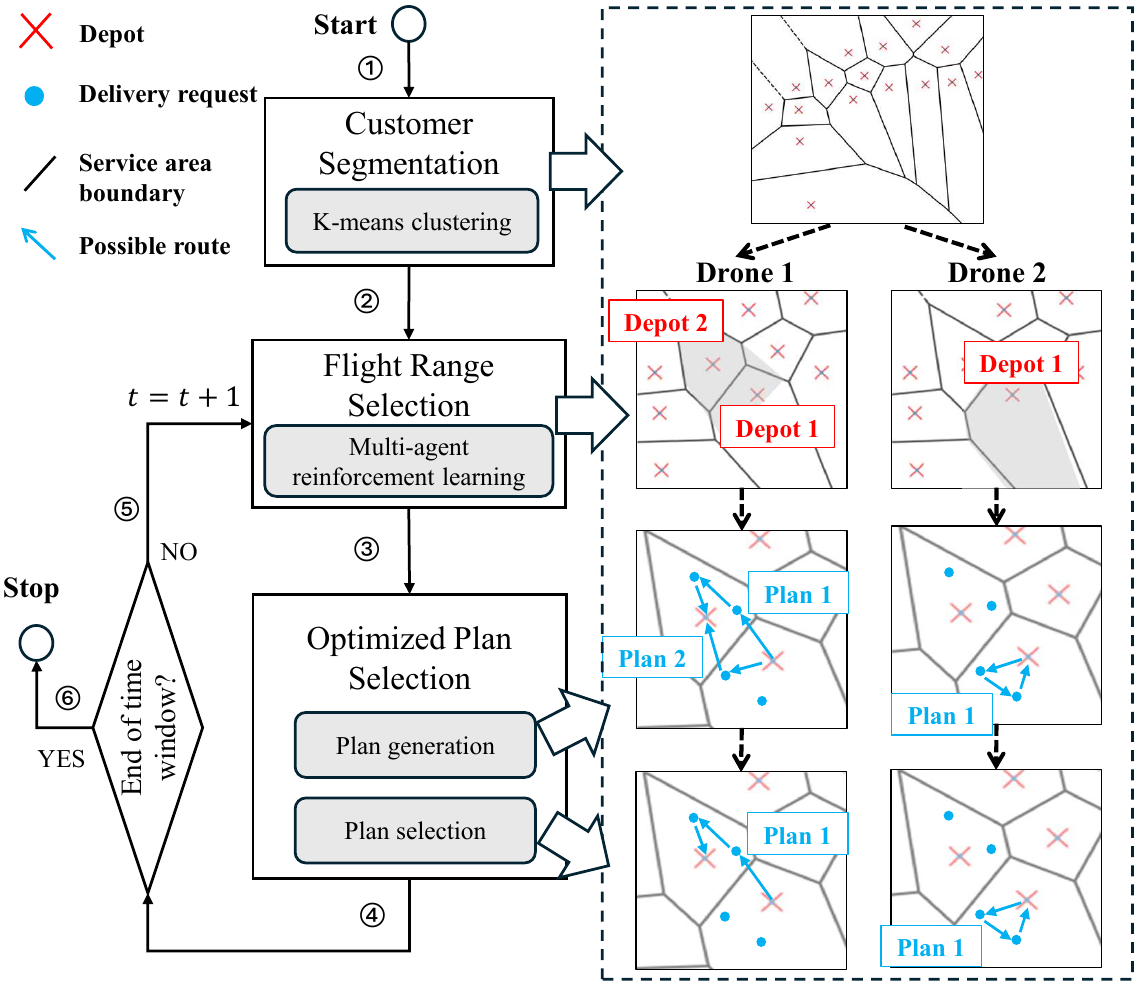}
    \caption{Overall framework of the methodology to solve multi-drone delivery problem. }
    \label{fig:framework}
\end{figure}

Fig.~\ref{fig:framework} illustrates the overall framework of the methodology. Firstly, the K-means clustering algorithm is used for customer requests segmentation, which partitions the map into a number of service areas where the depots are located at their centers. Then, the \emph{MAR-OPS} approach is introduced to select a flight range of drones via reinforcement learning and solves the optimized plan selection problem of drones within the selected flight range. In flight range selection, each drone determines a flying direction that identifies the depots as the start and end points for the route (e.g., from Depot 1 to Depot 2 for drone 1, and from Depot 1 to Depot 1 for drone 2). Along with depots, their corresponding service areas must be covered by the flight range of the drone. Under the flight range, drones serve the customer requests and generate all possible routes, i.e., the task plans. The MIP solver selects the optimized plans for execution. Finally, drones reselect a new flight range based on their current locations and the state of customer requests in the next time window.

\begin{algorithm}[t]
\SetAlgoLined
    \caption{K-Means Clustering Algorithm.}
    \label{algorithm1}
    \textbf{Input:} Customers dataset $\mathcal{D}$, number of depots $N$\;
    \While{depot locations do not change}
    {
        Randomly select $N$ centers of service areas as depot locations\;
        Calculate the Euclidean distances between depots and customers\;
        Divide $\mathcal{D}$ into $N$ clusters based on the distances\;
        Adjust the coordinates of depots to be the centers of corresponding clusters\;
    }
    Set the boundaries between clusters\;
    \textbf{Output:} Set of depot nodes $\mathcal{N}$, service areas $\mathcal{N}_a$
\end{algorithm}

\subsection{K-means Clustering Algorithm} \label{subsec:cluster}
Before determining the flight range, the service areas with centers (i.e., the locations of depot nodes) and boundaries are obtained, which is executed before drones performing delivery. The algorithm uses K-means clustering algorithm with a historical dataset of customers that contains the information of index, X/Y coordinates, and customers' demand time. This is because K-means clustering can group the historical customer requests that are geographically close. We also positions depots at cluster centroids, ensuring they are centrally located relative to customer requests. This decreases the travel distance from depot to customer requests, thereby reducing the energy consumption. Additionally, K-means clustering is more computationally efficient and easy to implement in the large delivery dataset (e.g., millions of delivery requests) compared to other clustering algorithms such as DBSCAN and hierarchical clustering.

As shown in Algorithm~\ref{algorithm1}, the proposed algorithm calculates and modifies the centers of clusters according to the Euclidean distances between the centers and customer until convergence, that is, until the centers do not change (Lines 3-6). Once the final cluster centroids are determined (serving as depot nodes), linear decision boundaries are constructed by drawing perpendicular bisectors between every pair of centroids. Each perpendicular bisector represents a boundary where all points along it are equidistant to the two corresponding depot nodes. As shown in Fig.~\ref{fig:framework}, these boundaries--highlighted as yellow grid lines--divide the service areas and are positioned between the depot nodes, which are shown as red points.

\begin{figure}
    \centering
    \includegraphics[width=0.7\linewidth]{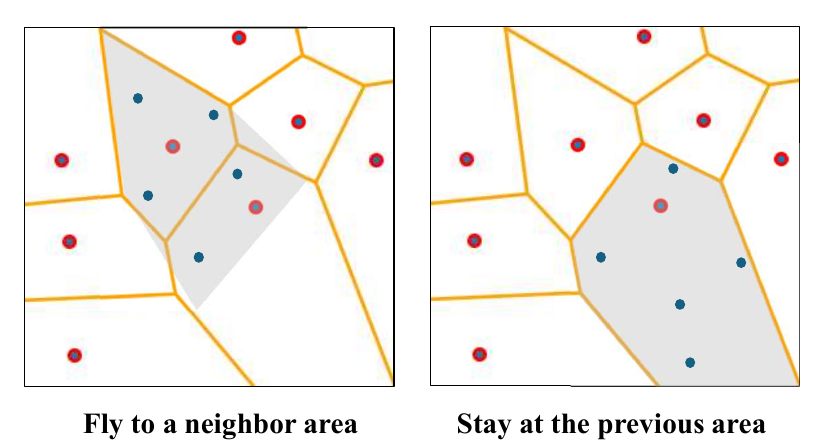}
    \caption{The flight range of a drone between its start and end depots (left is to select a neighbouring service area, right is to select current area). The yellow grids are the boundaries of service areas; grey shadow indicates the flight range; red points denote the depots; blue points denote the customers within the range that the drone can select. }
    \label{fig:fly_range}
\end{figure}

\subsection{Reinforcement Learning}
After finding the depot locations and the boundaries of service areas, each drone $u$ can strategically find its flight range $R_u$ by choosing a service area. As shown in Fig.~\ref{fig:fly_range}, a drone selects a neighboring depot or the depot at which it is currently locating as the destination, and then determines the flight range that covers the corresponding service areas. Once drones take actions to select service areas, they change their current locations and battery levels while observing the delivery delay of customers nearby. Therefore, the problem scenario can be represented as a Markov decision process. We model the problem using state, action, and reward concepts: 

1) \emph{State}: The state $s(t)$ at time window $t$ consists of four components $(\mathcal{S}_1, \mathcal{S}_2, \mathcal{S}_3, \mathcal{S}_4)$, where $\mathcal{S}_1$ represents the current location of drone $u$ within one of the service areas; $\mathcal{S}_2$ reflects the current battery level of $u$, derived from the energy consumption model; $\mathcal{S}_3 $ captures the delivery delay of customers within the service area where $u$ is currently located; $\mathcal{S}_4 $ records the delivery delay of customers in neighboring areas adjacent to the location of $u$. The set of customers observed by $u$ is denoted as $\mathcal{O}_u(t)$.

2) \emph{Action}: The action $a(t) = \{ a_1(t),...,a_U(t) \}$ at time window $t$ consists of the indexes of service areas selected by drones. They can choose one of neighboring service areas or stay at the previous area. Specifically, if the action has five values, then $a(t) = 0$ indicates that the drone returns to the start depot after delivering, whereas $a(t) = 1, 2, 3$ represents that the drone chooses one of three neighboring service areas.

3) \emph{Reward}: The reward $r_u(t)$ evaluates the flight range selection of the drone. Based on the objective function in Eq.(\ref{eq:objective_delay}) and (\ref{eq:objective_energy}), the expected immediate local reward of one drone at time window $t$ can be defined as the negative delivery delay and energy cost. Both two objective functions are normalized using Sigmoid. The function is formulated as: 
\begin{align}
    \begin{split}
        r_u(t) = -
        &(1 - \alpha) \sum_{i \in \mathcal{O}_u(t)} f_p (t, t_i)\\
        & - \alpha \sum_{i \in \mathcal{V}(t)} \sum_{j \in \mathcal{V}(t)} E_{uij} \cdot x_{uij} (t),
    \end{split}
    \label{eq:reward}
\end{align}
where $\alpha$ is a parameter weight that makes a trade-off between the total delay of observed customers by $u$ and the energy cost of $u$, $0 \leq \alpha \leq 1$. This is because these objectives often conflict: minimizing delay may require more energy-intensive operations in long route, while minimizing energy usage may neglect high-urgent customers. By adjusting the weights between 0 and 1, the operators in transportation systems can tune the reward function to prioritize or balance competing objectives based on the specific needs of the application. For example, the system can prioritize minimizing delivery delays by decreasing $\alpha$ to address urgent customer demands during a pandemic, or focus on reducing energy consumption and carbon emissions to meet net zero targets by increasing $\alpha$.

\begin{algorithm}[t]
	\caption{Optimized Plan Selection.} 
	\label{algorithm2}
	\KwIn{Parameters of drones $u$ used in Section~\ref{sec:energy_model}, the flight range $R_u$, the maximum number of parcels $\hat{M}$, $\forall u \in \mathcal{U}$.}
	Initialize a set of plans $\mathbb{P} = \emptyset$\ and the sets of selected customer requests $\hat{\mathcal{C}_u} = \emptyset$ for each drone\;
    Find $\hat{M}$ customer requests with highest delivery delay within $R_u$, and store them into $\hat{\mathcal{C}_u}$\;
    Add all sets to $\hat{\mathcal{C}} = \sum_{u \in \mathcal{U}} \hat{\mathcal{C}_u}$\;
    \For{$\forall u \in \mathcal{U}$}
    {
        Find $K$ combinations of customer requests from $\hat{\mathcal{C}_u}$ under the constraint (\ref{eq:constraint_weight})\;
        \For{each plan index $k := 1,...,K$}
    	{
    		Find the shortest path among customer requests and departure/destination depots via greedy algorithm\;
            Generate the plan $p_{uk}$ that includes the indexes of selected customer requests\;
    		Calculate the energy consumption $E_{uk}$ based upon the path as the cost of the plan\;
    		Add the plan and cost to $\widehat{\mathbb{P}}$ \;
    	}
    }
    Set the objective function: $\min \; \sum_{u \in \mathcal{U}} \sum_{k \leq K} E_{uk} \cdot x_{uk}$\;
    Set the constraint (1): $ \sum_{k \leq K} x_{uk} = 1, \forall u \in \mathcal{U} $\;
    Set the constraint (2): $ \sum_{u \in \mathcal{U}} \sum_{k \leq K} \mathbf{1}\{c \in p_{uk}\} \cdot x_{uk} = 1, \forall c \in \hat{\mathcal{C}} $\;
    Minimize the objective function using an optimizer tool under constraints and find  the optimal route for each drone $p_u := p_{uk}$\;
	\KwOut{Optimal route $p_u$ for each drone.}
\end{algorithm}

\subsection{Route Planning Selection}\label{sec:branch&bound}
After selecting flight ranges $R_u$, $u \in \mathcal{U}$, drones run the optimized plan selection algorithm to generate and select optimized task plan for delivery. The depots and customer locations within the designated service areas are modeled as a graph, where each node represents a possible stop for the drones. This algorithm contains two parts: the plan generation and plan selection, see Algorithm~\ref{algorithm2}.

The plan generation part of the algorithm (Line 2-12) aims to enable each drone to generate task plans that cover the customer requests with the highest delivery delay within its flight range. Firstly, for the flight range $R_u$ of each drone $u$, $\forall u \in \mathcal{U}$, the algorithm searches for a number of customer requests with the highest delivery delay within $R_u$ and store them into a set $\hat{\mathcal{C}_u}$ (Line 2). This number is determined by the maximum number of parcels for each drone $\hat{M}$, which is empirically set according to the average weight of parcels and the payload of drones. Next, each drone $u$ finds all combinations (equal to $K$) of customer requests from its corresponding $\hat{\mathcal{C}_u}$ (Line 5). The combination denotes a number of requests ($\leq \hat{M}$) selected from $\hat{\mathcal{C}_u}$ whereas their total weight of parcels is no higher than the drone payload $M_u$. Then, each drone finds the shortest path for each combination of customer requests, calculates the energy consumption $E_{uk}$, and generate the plan $p_{uk}$ (Line 6-11). All plans and their corresponding cost (i.e., energy consumption) are stored into the set $\mathbb{P}$.

In the plan selection part (Line 13-16), the objective function is set to minimize the total cost of plans selected by drones, where $x_{uk}$ denotes binary variable which takes 1 if drone $u$ selects the plan $l$ and $0$ otherwise. Two constraints are then applied: (1) each drone must select exactly one plan, and (2) each customer request must be visited exactly once across all selected plans, where $\mathbf{1}\{c \in p_{uk}\}$ is an indicator function which takes 1 if $c \in p_{uk}$ and 0 otherwise. Once the constraints are set, the optimal route for each drone is then obtained, yielding the most cost-effective delivery plan. Finally, the algorithm selects an optimal route for each drone using an optimizer tool such as Gurobi~\cite{gurobi_2024}.

\begin{algorithm}[t]
    \caption{Training of \emph{MAR-OPS}.}
    \label{algorithm3}
    Initialize critic network $Q(\cdot)$ and actor network $\pi(\cdot)$ with parameters $\theta^Q$, $\theta^\pi$;
    Set the depot nodes and the boundaries of service areas (see Algorithm~\ref{algorithm1})\;
    \For{episode $:= 1$ to max-episode-number}
    {
        Reset environment and obtain the initial state\;
        \For{epoch $t := 1$ to max-episode-length}
        {
            \For{$\forall u \in \mathcal{U}$}
            {
                Take action $a_u(t) = \pi(\mathcal{O}_u(t) | \theta^\pi)$\;
                Find the destination depot and the flight range $R_u$ \;
            }
            Generate task plans and find the optimal one for each drone via Algorithm~\ref{algorithm2}\;
            \For{$\forall u \in \mathcal{U}$}
            {
                Compute reward and update state\;
                Store transition sample into buffer\;
                Sample a random mini-batch from buffer\;
            }
        }
        Estimate advantage via Eq.(\ref{eq:advantage})\;
        Calculate the probability ratio via Eq.(\ref{eq:prob})\;
        Update $\theta^\pi$ by minimizing the loss via Eq.(\ref{eq:clip_obj}) (\ref{eq:actor})\;
        Update $\theta^Q$ by minimizing the loss via Eq.(\ref{eq:critic})\;
    }
\end{algorithm}

\subsection{Proposed Algorithm Details}
Throughout the designed methodology, the actions of flight range selection play a pivotal role in obtaining high reward,  i.e., low delivery delay and energy consumption. A ``good'' action balances the immediate efficiency of the flight range with the potential to serve future customers effectively, ensuring minimal delay and optimal energy usage over time. For instance, as shown in Fig.~\ref{fig:scenario}, a ``good'' action might direct a drone to fly to Area 2 at $t$, where delays are currently high. This proactive choice positions the drone to serve delayed customers more efficiently in Area 4 at $t+1$, reducing future travel. However, a ``bad'' action may suggest the drone remain in Area 1 to minimize immediate energy use, but at the cost of greater delays later. Thus, the proposed algorithm \emph{MAR-OPS} leverages multi-agent deep reinforcement learning to reinforce ``good'' actions and lower the probability to choose ``bad'' actions. It learns an optimal policy by exploring various choices of flight range, adapting to the dynamic environment and evaluating their outcomes over time.

The algorithm learns the policy using an extension of the actor-critic policy gradient approach, employing two deep neural networks for each agent: a critic network $Q(\cdot)$, and multiple actor networks $\pi(\cdot)$. The actor network provides actions for agents and the critic network evaluates the Q-value of these actions. In addition, \emph{MAR-OPS} employs Proximal Policy Optimization (PPO)~\cite{jiang2023underwater} to prevent detrimental updates and improving the stability of the learning process.

The critic network estimates the reward associated with a transition using the Bellman equation. Its parameter $\theta^Q$ is updated by minimizing a loss function based on an advantage function $A_{uk}$. The advantage function is expressed as follows:
\begin{equation}
    A_{uh} = r_{uh} + \gamma \cdot Q(o_{u, h+1}, a_{u, h+1}) - Q(o_{uh}, a_{uh}),
    \label{eq:advantage}
\end{equation}
where $\gamma$ is a discount factor; $o_{uh}$ and $a_{uh}$ denote the observation and the action of agent $u$ in the sample $h$ in a mini-batch from buffer, $h \leq H$. 

Then, the critic network provides $A_{uh}$ to the actor network to increase the probability of actions that have a positive impact and decrease the ones that have negative impact. The actions are taken by drones as $a_u(t) = \pi(o_u(t) | \theta^\pi)$. To update the parameter $\theta^\pi$ of the actor network, PPO utilizes a policy ratio $\text{prob}(\theta^\pi, u)$, which is formulated as:
\begin{equation}
   \text{prob}(\theta^\pi, u) = \frac{\pi_{\theta^\pi}(a_{uh} | o_{uh})}{\pi_{old}(a_{uh} | o_{uh})},
   \label{eq:prob}
\end{equation}
where $\pi_{old}$ denotes the older policy of the actor network in the previous iteration. This policy ratio is used to calculate the clip surrogate objective $s_{uh}$:
\begin{align}
    \begin{split}
        s_{uh} = \min [ \;
        & \text{prob}(\theta^\pi, u) \cdot A_{uh}, \\
        & \text{clip}(\text{prob}(\theta^\pi, u), 1 - \epsilon, 1 + \epsilon) \cdot A_{uh} 
        ],
    \end{split}
    \label{eq:clip_obj}
\end{align}
where $\epsilon$ is a hyperparameter; $clip(\cdot)$ indicates the clipping method to restrict the range of $ratio(\theta^\pi, u)$ in order to preventing incentives from exceeding the interval $[1 - \epsilon, 1 + \epsilon]$. Finally, the parameters of both actor $\theta^\pi$ and critic $\theta^Q$ networks are updated by minimizing the loss functions $L_{\text{critic}}(\theta^Q)$ and $L_{\text{actor}}(\theta^\pi)$ respectively. They are expressed as follows:
\begin{equation}
    L_{\text{actor}}(\theta^\pi) = \frac{1}{U \cdot H}  \sum_{u=1}^{U} \sum_{h=1}^{H} s_{uh},
    \label{eq:actor}
\end{equation}
\begin{equation}
    L_{\text{critic}}(\theta^Q) = \frac{1}{U \cdot H} \sum^U_{u=1} \sum^H_{h=1} (A_{uh})^2.
    \label{eq:critic}
\end{equation}

Therefore, the overall training process of the methodology primarily involves the following steps (see Algorithm~\ref{algorithm3}): At the beginning of each episode, the environment is reset with the input of delivery data, which includes the coordinates and time of customer requests. The state of drones are also initialized. Next, the algorithm runs the K-means clustering in Algorithm~\ref{algorithm1} for customer requests segmentation. Then, each agent (or drone) takes an action to select a service area based on its current state. These actions also determine the departure and destination of drones, i.e., the start and end depots. After generating and selecting a task plan for delivery through Algorithm~\ref{algorithm2}, drones calculate their immediate reward and transition to a new state. The buffer, a data storage structure used for experience replay, stores all transitions of each drone. Several groups of transitions are sampled randomly ($H$ groups of transitions) for updating the parameters of both the critic and actor networks. Finally, the algorithm updates the parameters in critic network and actor network.

\begin{figure}
    \centering
    \includegraphics[width=\linewidth]{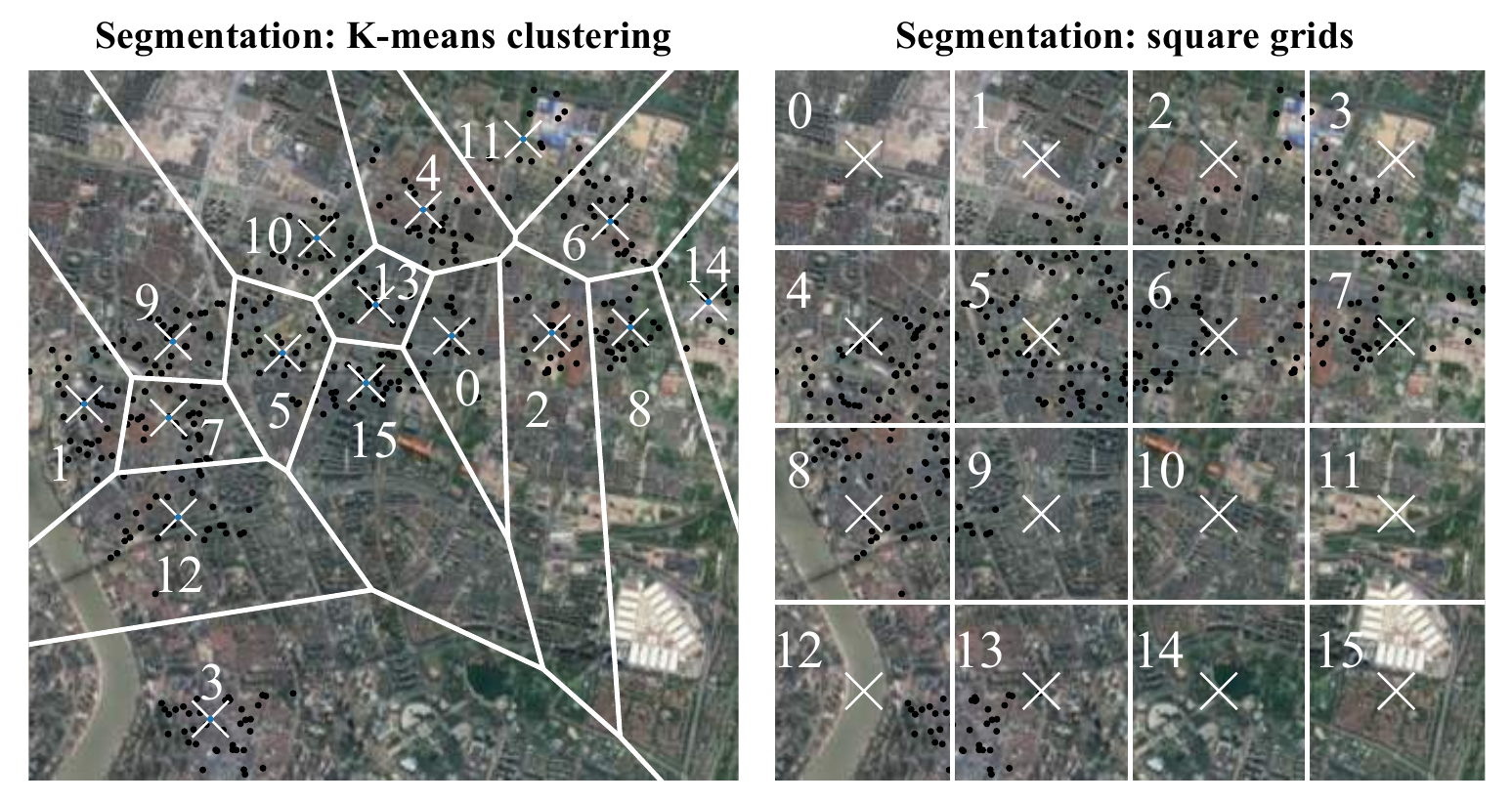}
    \caption{A city map with the area of $10 \times 10km$ in Shanghai City. The left picture illustrates an example of the distribution of customer demands on August 20, 2022 (shown in yellow points). The right one shows the distribution of depots (orange points) and the boundaries of service areas (red lines) determined by clustering based on LaDe dataset.}
    \label{fig:city_map}
\end{figure}

\begin{table}  
	\caption{Notations for drones.}  
	\centering
	\begin{tabularx}{7cm}{lll}  
		\hline  
		Notation & Description & Value \\  
		\hline    
		$ m_{b} $ & mass of drone body & $2 \; kg$  \\
		$ m_{e} $ & mass of battery & $1 \;kg$  \\
		$ d $ & diameter of propellers & $0.5 \; m$  \\
		$ r $ & number of  propellers & $4$  \\
		$ v $ & ground speed & $10 \; m/s$  \\
		$ \epsilon_u $ & power efficiency & $0.8$  \\
		\hline  
	\end{tabularx}  
	\label{table:drone}
\end{table}

\section{Performance Evaluation \label{sec:results}}
In this section, an overview of the experimental settings is presented. The delivery dataset, specification of drones, and the used neural networks are introduced. Then, the baselines and performance evaluation metrics are discussed. Finally, the results are assessed across various scenarios.

\subsection{Experimental Settings}
\cparagraph{Delivery dataset and scenarios} In order to evaluate \emph{MAR-OPS}, we model a real-world delivery scenario in Shanghai City based on a last-mile delivery dataset, named LaDe~\cite{wu2023lade}. LaDe is a large-scale dataset consisting of data, such as time and location, for 10,677,000 packages collected by 21,000 couriers in 5 cities across 6 months. The dataset comprises data for both pickup and delivery of packages for 30 regions in Shanghai City. 

\cparagraph{Scenarios}
As shown in Fig.~\ref{fig:city_map}, this study selects an area of $10 \times 10km$ in the city where a number of customers request parcels at different times from 9 am to 5 pm. It illustrates an example of the distribution of customer requests on August 20, 2022 (shown in black points). The left picture shows the distribution of 16 depots (white crosses) and the boundaries of service areas (white lines) with indexes (white number) determined by K-means clustering based on LaDe dataset. The right one shows the depots and boundaries determined by square grids, which will be used for baseline comparison in Section~\ref{sec:baseline}. The total duration is divided into $12$ time windows, each representing $30$ min, when a drone executes a delivery mission and fly back to a depot. Except for X/Y coordinates, each request has an actual delivery time in the dataset, which is assumed to be the expected delivery time. The cross-validation is employed for $90$ days of data: $80\%$ for training and $20\%$ for testing.


\cparagraph{Drones} This paper refers to the parameters of Meituan drone because of its high flight range and payload~\cite{meituan2023}. Each drone has a maximum flight range of around $3$ km and a maximum payload of around $2.5$ kg. The other drone parameters used in this paper for the simulations are set as listed in the Table~\ref{table:drone}.

\cparagraph{Neural network and learning algorithm}
A total of $H = 64$ groups of transitions are sampled as mini-batches in a replay buffer, with a discount factor of $\gamma=0.95$ and a clip interval hyperparameter of $0.2$ for policy updating. We use the recurrent neural network, and use $64$ neurons in the two hidden layers of the recurrent neural network in both critic and actor networks. The activation function used for the networks is tanh. The models are trained over $\mathcal{E}=10,000$ episodes, each consisting of multiple epochs (equal to the number of time windows).

\subsection{Baselines and Metrics} \label{sec:baseline}
A direct comparison between the proposed methods and existing work is challenging due to the limited relevant work. Most existing approaches are not well-suited for multi-objective combinatorial optimization problems that involve both time-sensitive constraints and multi-parcel delivery requirements. For this reason, this paper compares the proposed method \emph{MAR-OPS} with the following relevant methods:
\begin{itemize}
    \item \emph{OPS-global}: The flight range of each drone is set as the default, that is, every drone searches the customer requests over the whole map within each time window regardless of its flight range. It does not solve customer requests segmentation and flight range selection.
    \item \emph{OPS-random}: K-means clustering algorithm is leveraged to determine the service areas; drones select their sub-ranges randomly. \emph{OPS-random} does not support any long-term strategic navigation between depots.
    \item \emph{MAPPO}: It is the state-of-the-art MARL algorithm using proximal policy optimization~\cite{jiang2023underwater}. Different from \emph{MAR-OPS} that takes actions to select a flight range, the action of this method is to choose a task plan from the set $\hat{\mathcal{C}_u}$ without the help of Gurobi optimizer.
    \item \emph{Sqaures}: It leverages MARL-based flight range selection and optimized plan selection algorithms, but divides the map into square grids as service areas for customer requests segmentation instead of K-means clustering. 
\end{itemize}

All the algorithms are evaluated using the following two key metrics: (1) \emph{Mean energy consumption}, which denotes the mean energy consumption per drone that travels within all time window based on the objective function (\ref{eq:objective_energy}); (2) \emph{Average delivery delay}, which indicates the average delay time (hours) per customer request all time windows based on the objective function (\ref{eq:objective_delay});

Moreover, several other metrics can also be observed on these algorithms: (1) \emph{Combined cost}, which is the sum of normalized mean energy consumption and average delivery delay. (2) \emph{Delay unfairness}, which measures the delivery delay inequality among search areas. Specifically, the total delivery delay of customer requests in each service area is calculated, and the unfairness value between these total delivery delay values is calculated using Gini coefficient. (3) \emph{Average running time}, which calculates the average time (seconds) it takes these algorithms to run per repetition. (4) \emph{Average early arrival time}, which measures the average amount of time (hours) by which drones arrive earlier than the expected delivery time (or deadline) across all customer requests over all time windows. (5) \emph{Depot load}, which quantifies the total weight (kg) of the parcels picked up by the drones from each depot over all time windows.

\newcommand{\tabincell}[2]{\begin{tabular}{@{}#1@{}}#2\end{tabular}}
\begin{table*}[!t]
\footnotesize
	\caption{Performance comparison of five methods on the basic delivery scenario.}  
	\centering
    \footnotesize
    \resizebox{\linewidth}{!}
    {
    	\begin{tabular}{llllll}  
		\toprule  
        \textbf{Metrics} & \emph{OPS-global} & \emph{OPS-random} & \emph{MAPPO} & \emph{Squares} & \emph{MAR-OPS}\\
        
		\midrule    
		Mean energy consumption (kJ) & $1551 \pm 106$ & $662.0 \pm 40.4$ & $664.3 \pm 33.2$ & $1768 \pm 126.7$ & $669.1 \pm 33.2$  \\
        Average delivery delay (hour) & $0.67 \pm 0.20$ & $1.17 \pm 0.16$ & $1.46 \pm 0.10$ & $1.15 \pm 0.11$ & $0.88 \pm 0.10$ \\
        Combined cost & $0.66 \pm 0.09$ & $0.57 \pm 0.06$ & $0.67 \pm 0.04$ & $0.87 \pm 0.07$ & $\textbf{0.48} \pm \textbf{0.04}$ \\
		Delay unfairness & $0.21 \pm 0.01$ & $0.60 \pm 0.02$ & $0.34 \pm 0.02$ & $0.56 \pm 0.08$ & $0.41 \pm 0.01$ \\
		Average running time (second) & $6.23 \pm 0.11$ & $0.28 \pm 0.04$ & $0.38 \pm 0.02$ & $0.46 \pm 0.02$ & $0.46 \pm 0.02$   \\
        Average earlier arrival time (hour) & $0.03 \pm 0.06$ & $0.52 \pm 0.17$ & $1.03 \pm 0.05$ & $0.11 \pm 0.04$ & $0.17 \pm 0.06$\\
		\bottomrule  
	\end{tabular}
    }
	\label{table:basic}
\end{table*} 

\begin{figure}[!t]
    \centering
    \includegraphics[width=\linewidth]{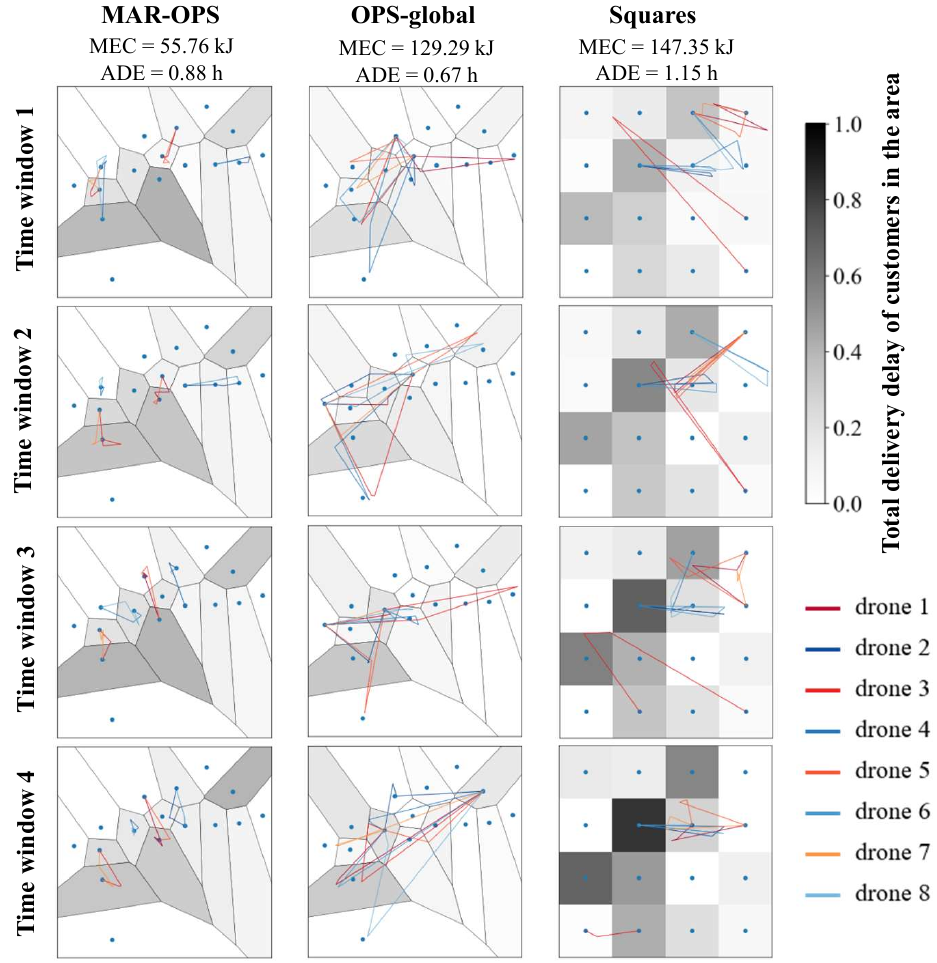}
    \caption{Comparison of optimal drone trajectories among all three approaches in
different time windows (MEC is mean energy consumption, ADE is average delivery delay).}
    \label{fig:trajectories}
\end{figure}

\subsection{Results and Analysis}
\begin{figure}[!t]
    \centering
    \subfigure[Trade-off weight vs. delivery delay \& average energy cost.]{
        \includegraphics[width=0.47\linewidth]{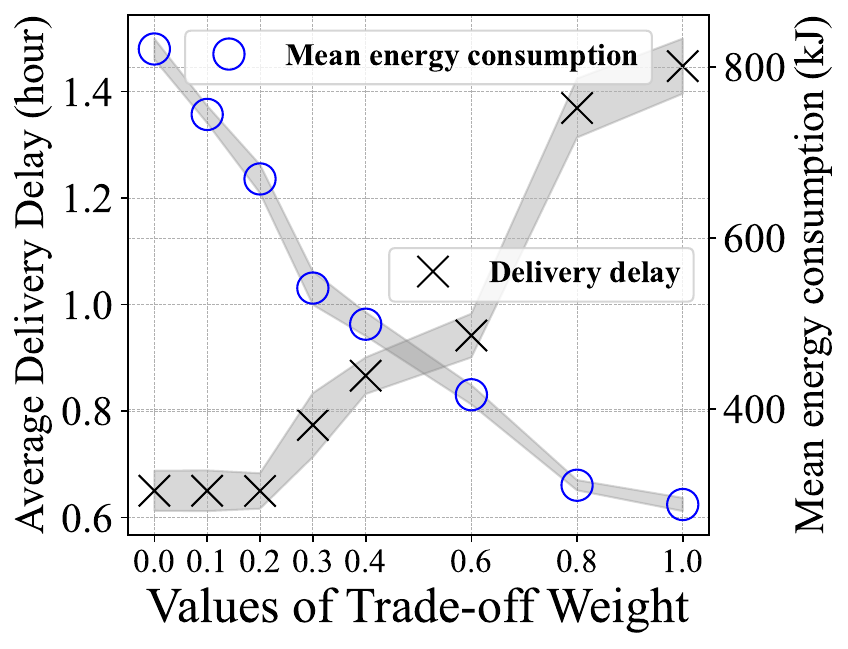}
        \label{fig:tradeoff_basic}
	}
    \subfigure[Trade-off weight vs. combined cost.]{
		\includegraphics[width=0.40\linewidth]{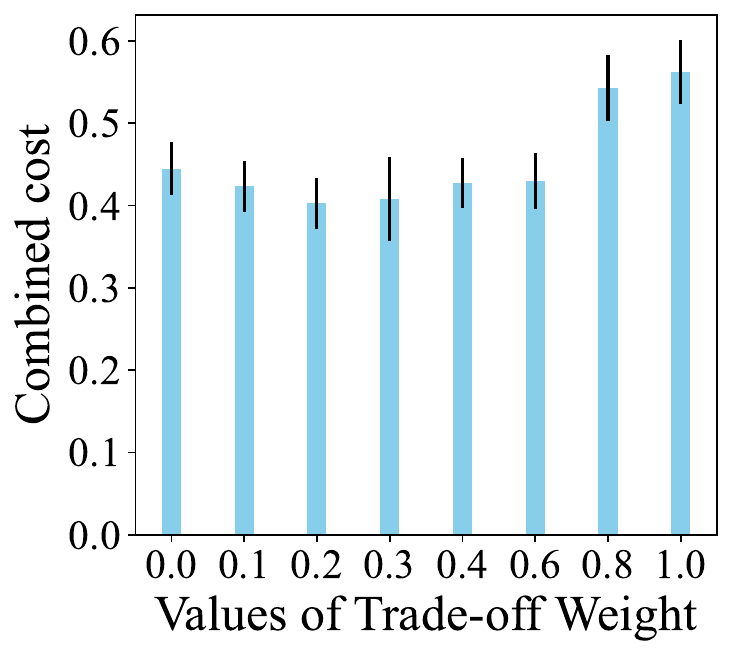}
		\label{fig:tradeoff_combined}
	}
    \caption{Performance comparison of \emph{MAR-OPS} across different values of trade-off weight. The shadow represents the error (i.e., standard deviation) of metrics. \textbf{\emph{MAR-OPS} can achieve either low mean energy consumption or low average delivery delay by adjusting the tradeoff parameter. The high cost of energy consumption comes with benefits on timely delivery and vice versa.}}
    \label{fig:tradeoff}
\end{figure}

\begin{figure}[!t]
    \centering
    \includegraphics[width=0.9\linewidth]{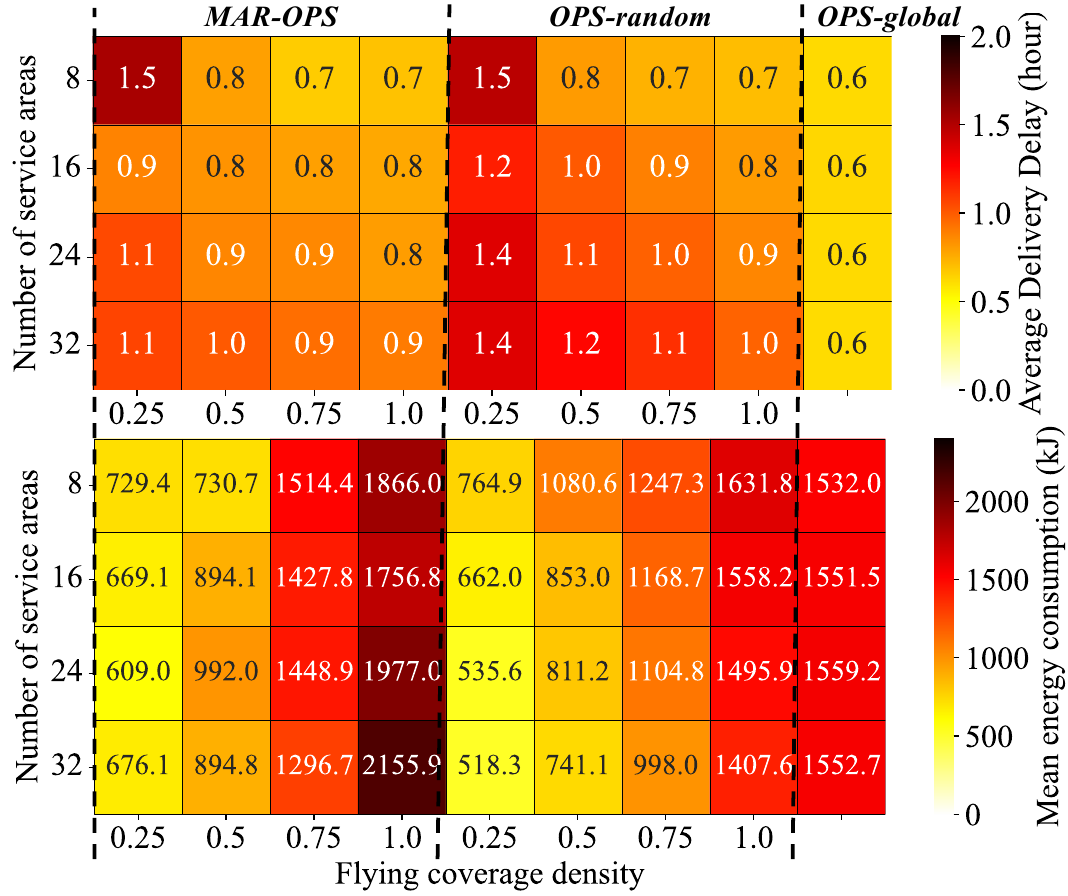}
    \caption{Performance comparison of three approaches across different flying coverage density and number of service areas. \textbf{\emph{MAR-OPS} searches customers and reduces delivery delay more energy-efficiently than \emph{OPS-random}, which operates with a low flying coverage density.}}
    \label{fig:depots}
\end{figure}

The \emph{basic delivery scenario} is evaluated at first. There are $16$ service areas with $8$ dispatched drones. The number of actions is $4$, which indicates that each drone selects one of the four nearest depots as the destination (including its departure depot). The tradeoff parameter is set as 0.5 as default. The parameters are set according to the optimal performance of \emph{MAR-OPS} (see Fig.~\ref{fig:tradeoff} and Fig.~\ref{fig:depots}). The purpose is to compare the performance of different approaches. 

Table~\ref{table:basic} shows the metric performance of all methods which run $40$ times to obtain average values and standard deviation errors. Among all methods, \emph{OPS-global} has the lowest average delivery delay and delay unfairness, but incurs the highest average running time due to its extensive search range. When the search range is reduced, \emph{MAR-OPS} can run approximately $93.73\%$ faster than \emph{OPS-global}, making it more suitable to real-time application. \emph{OPS-global} also has the higher mean energy consumption than \emph{MAR-OPS} by $882.4$kJ since drones are more likely to choose distant customers regardless of their maximum flying time. The comparison of drone trajectories is illustrated in Fig.~\ref{fig:trajectories}. Each drone takes around $80.80\%$ of battery capacity on average, exceeding its safe battery level. Based on the average carbon intensity in United Kingdom in 2024, which is $125g$ $CO_2$ per $kWh$~\cite{edenseven2025}, \emph{MAR-OPS} can reduce approximately $245$g $CO_2$ less than \emph{OPS-global}. In overall, \emph{MAR-OPS} has $27.27\%$ lower combined cost than \emph{OPS-global}, validating the energy-efficiency for delivery service.

Except for \emph{OPS-global}, \emph{MAR-OPS} has lower delivery delay than other methods, by controlling drones to strategically target high-delay areas while still maintaining low energy consumption. While \emph{OPS-random} minimizes the mean energy consumption, it lacks a long-term strategy that guides drones to the customer requests with the highest delivery delay. As a result, drones using \emph{OPS-random} arrive earlier than those using \emph{MAR-OPS} by an average of $0.35$ hour, but they also experience longer delays, averaging $0.29 $ hour more. This random selection of flight range also leads to unequal serving of customer requests, with $0.19$ higher than \emph{MAR-OPS}. Even though \emph{MAPPO} prevents drones from choosing incorrect service areas, it performs the worst in terms of delivery delay, with the highest average delivery delay and earlier arrival time. This is due to inefficient exploration in training caused by high action space. In addition, \emph{MAR-OPS} using K-means clustering significantly outperforms \emph{Squares} using square grids with a $1099$kJ reduction in energy and $0.27$ hour decrease in delay. This is because the K-means clustering restrict drones to fly over high request density areas, whereas square grids may result in that drones waste their energy to choose a corner grid without any delivery requests, see Fig.~\ref{fig:trajectories}.

The \emph{complex delivery scenario} is then studied by varying three parameters: (1) The trade-off weight to balance the delivery delay and average energy cost defined in the reward function; (2) the number of drones used in the delivery mission; and (3) the flying coverage density to represent the ratio of the number of areas a drone chooses to fly over the total number of service areas in the map. The goal is to assess the scalability of the proposed approaches under different experimental conditions. 

\cparagraph{Trade-off weight} Fig.~\ref{fig:tradeoff_basic} illustrates the influence of trade-off weight on the balance between mean energy consumption and average delivery delay of \emph{MAR-OPS}. The shadow represents the error (i.e., standard deviation) of metrics. As the value of trade-off weight $\alpha$ increases, the energy consumption of each drone decreases linearly while the delivery delay of customers increases significantly when $\alpha > 0.2$. The heavier trade-off weight on the energy cost restricts drones to travel less. As a result, drones fail to deliver distant customers request with high delay. When $\alpha \geq 0.6$, drones only pick-up parcels from area $0$, $5$, $13$ and $15$, see Fig.~\ref{fig:city_map}. \emph{MAR-OPS} only produces $80.04g$ $CO_2$ at the cost of $1.45$ hour average delivery delay when $\alpha = 1$. In contrast, lower $\alpha$ reduces the delivery delay by controlling drones to fly across all areas and take parcels from their corresponding depots. \emph{MAR-OPS} achieves a relatively equal average delivery delay with \emph{OPS-global} when $\alpha = 0$, but $730.3$kJ lower mean energy consumption. In overall, see Fig.~\ref{fig:tradeoff_combined}, \emph{MAR-OPS} has the lowest combined cost when $\alpha = 0.2$ (chosen for basic delivery scenario), which represents the most balanced operation point where the drone delivery service runs energy-efficiently and timely.

\begin{figure}[!t]
    \centering
    \subfigure{
        \includegraphics[width=0.45\linewidth]{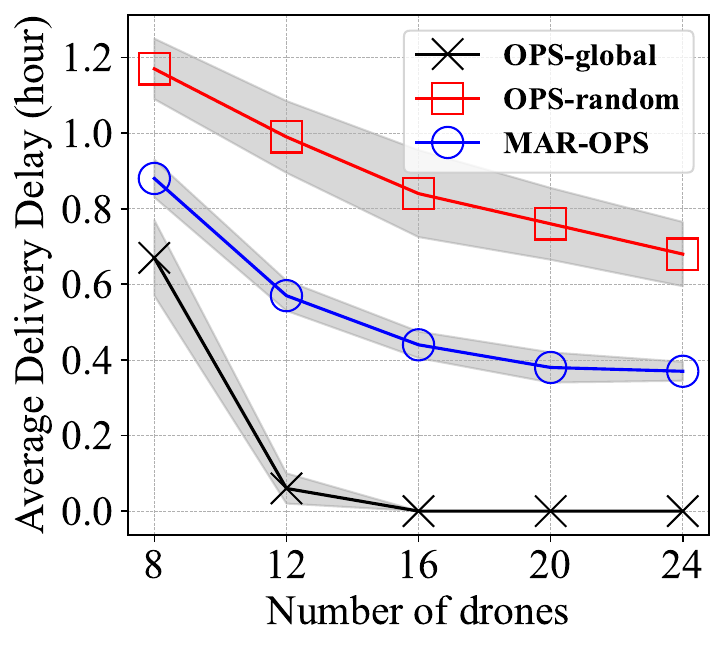}
	}
    \subfigure{
		\includegraphics[width=0.45\linewidth]{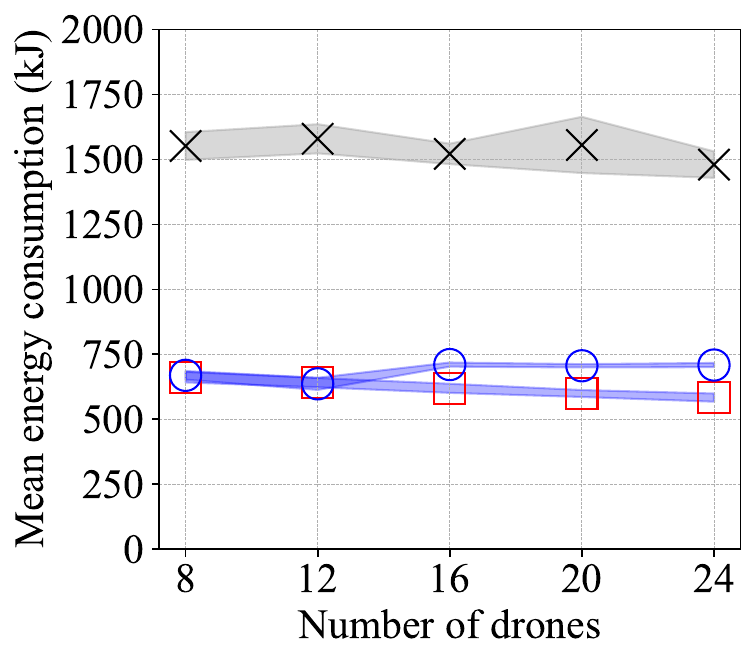}
	}
    \subfigure{
        \includegraphics[width=0.45\linewidth]{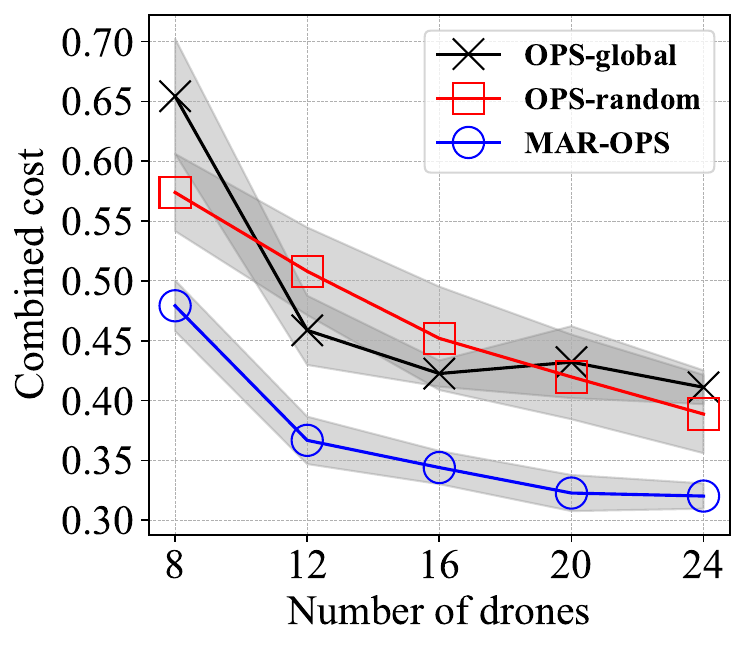}
	}
     \subfigure{
        \includegraphics[width=0.45\linewidth]{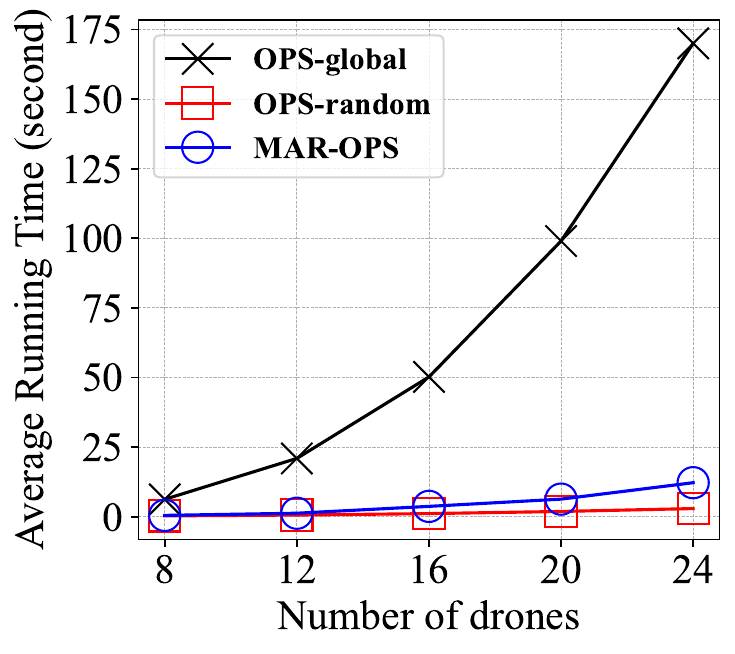}
	}
    \subfigure{
        \includegraphics[width=0.7\linewidth]{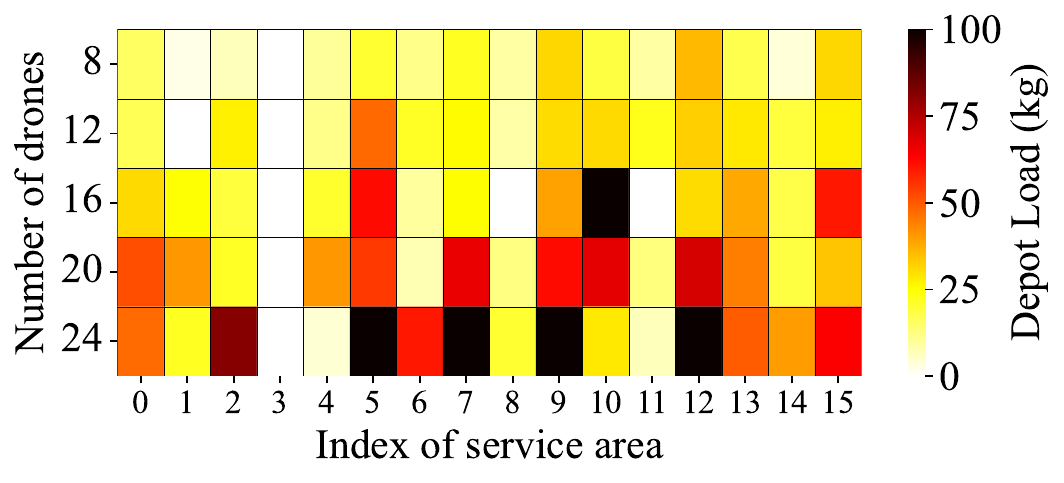}
	}
    \caption{Performance comparison of three methods in average delivery delay, mean energy consumption, combined cost, average running time, and depot load across different number of drones. \textbf{High number of drones decreases average delivery delay of \emph{MAR-OPS}, while maintaining a low mean energy consumption and a low running time.}}
    \label{fig:drones}
\end{figure}


\cparagraph{Number of drones} As shown in Fig.~\ref{fig:drones}, if the number of drones increases, the average delivery delay of all methods decreases as drones can cover a wider area and deliver to customer requests on time. This proves that sufficient drone resources can minimize the arriving time of all customer deliveries and efficiently mitigate the customer delay. The effect is significant in \emph{OPS-global} which has has $0$ delay with more than $16$ drones and the lowest delay unfairness, albeit with high energy consumption and running time. When the number increases to $20$, \emph{MAR-OPS} has the lower mean energy consumption by approximately $850.0$kJ and average running time by around $88.8$ seconds compared to \emph{OPS-global}, leading to $21.95\%$ lower combined cost. This demonstrates the applicability of the proposed approach in energy-constrained real-time delivery scenarios. Moreover, \emph{MAR-OPS} learns the flying strategy efficiently, decreasing the delivery delay by approximately $24.25\%$ compared to \emph{OPS-random}. In terms of the depot load, drones are gradually required to take parcels from all except the depots in service areas $3$ and $11$ (see Fig.~\ref{fig:city_map}), as the number of drones increases. This is because these low-load depots are located at the borders of the map and are hard to access.


\cparagraph{Flying coverage density} When the flying coverage density increases from $0.25$ to $1.0$, the number of service areas a drone chooses to fly increases by fixing the total number of service areas. Fig.~\ref{fig:depots} illustrates the performance of \emph{MAR-OPS} and baseline methods when using varying flying coverage densities and numbers of service areas. On the one hand, if the flying coverage density increases from $0.25$ to $1.0$, the flight range of both \emph{MAR-OPS} and \emph{OPS-random} extends, which helps drones to reach customer requests with higher delivery delays, thereby reducing delivery delay of the system. As the coverage density rises, the mean energy consumption increases due to a greater likelihood to choose longer flight routes. Note that in the case of high flying coverage densities, \emph{MAR-OPS} typically selects only a single flight range, rather than covering the entire map. This narrower range can make it more challenging to find optimal routes compared to \emph{OPS-global}. On the other hand, if the number of service areas increases from $16$ to $32$, while fixing flying coverage density, \emph{MAR-OPS} maintains mean energy consumption with only an $8.96\%$ variance but achieves a $12.52\%$ reduction in average delivery delay compared to \emph{OPS-random}. It is important to note that too few coverage options can hinder learning efficiency, as seen in the scenario with $8$ service areas at $0.25$ density.


In summary, \textbf{several new insights} are obtained from experimental results that are summarized as follows: (1) The proposed optimized plan selection (\emph{OPS}) algorithm effectively coordinates energy-aware drones to minimize delivery delays, while supporting energy-efficient multi-parcel delivery. (2) Strategic selection of flight range using reinforcement learning and clustering algorithms in \emph{MAR-OPS} ensures low delivery delay and effectively prevents drones from searching high distant customer requests. It not only reduces energy consumption and carbon emissions of drones, but enhances operation speed, making it highly adaptive to real-time delivery applications. (3) The tradeoff between energy consumption of drones and delivery delay of customer requests allows operators in transportation systems to make appropriate adjustments according to their priorities, such as net zero sustainability and pandemic delivery. (4) Experimental results on depot deployment optimization, addressing aspects such as the number, location, and parcel load distribution of depots, support service providers with infrastructure planning, thereby enhancing overall logistics efficiency and adaptability to dynamic customer requests.

\section{Conclusion and Future Work \label{sec:conclusions}}
This paper studies the last-mile delivery problem by a swarm of drones, which aims to minimize both energy consumption of drones and delivery delay of customer requests. Due to the constraints of flight range and parcel weight, the problem is modeled by a multi-objective combinatorial optimization problem using a multi-parcel energy consumption model. This problem is then decomposed into three sub-problems, which are addressed by three approaches respectively: (1) The clustering algorithm for customer requests segmentation; (2) The reinforcement learning approach for flight range selection; (3) The optimized plan selection algorithm for delivery navigation. A synthesis of these approaches is proposed to improve the long-term delivery efficiency by learning the policy on selecting an optimal flight range using \emph{MAR-OPS}. Extensive experimentation using a real-world delivery dataset from Shanghai city provides valuable insights for sustainable, efficient drone delivery. The results demonstrate that \emph{MAR-OPS} not only makes a tradeoff and minimizes both energy consumption (or carbon emissions) and delivery delay, but shows rapid operational speed and recommends depot deployment, making it adaptive for real-world logistics.

In the future work, a large number heterogeneous drones with different batteries, payloads, and power efficiencies will be considered. This will aim to support the large-scale delivery that requires the decentralized combinatorial optimization approaches, such as collective learning~\cite{narayanan2024large,pournaras2018decentralized}. Furthermore, this work needs to conduct experiments on a real-time drone testbed~\cite{qin20223,10639825} to validate the applicability of the proposed method.

\section*{ACKNOWLEDGEMENTS}
This research is supported by a UKRI Future Leaders Fellowship (MR-/W009560/1): 
\emph{Digitally Assisted Collective Governance of Smart City Commons–ARTIO}, and EPS International Academic Mobility Pump-priming Fund. This research is also partly supported by EU MSCA project COALESCE (n.101130739), and Research Council of Finland via X-SDEN (n.349965) and ECO-NEWS (n.358928).

\bibliographystyle{IEEEtran}
\bibliography{ref.bib}

\end{document}